\documentclass[10pt, twocolumn]{article}  % Two-column format

\usepackage[utf8]{inputenc}         % Handles special characters
\usepackage{amsmath, amssymb}       % Math symbols and formulas
\usepackage{graphicx}               % For including images
\usepackage{cite}                   % For managing references
\usepackage{hyperref}               % For clickable links
\usepackage{geometry}               % Adjust page layout
\usepackage{fancyhdr}               % Header and footer customization
\usepackage{titlesec}               % Control section headings
\usepackage{subfig}
\usepackage{subcaption}
\usepackage{wrapfig}
\usepackage{float}
\usepackage{hyperref}
\usepackage{parskip}

\title{\textbf{Occlusion aware obstacle prediction using people as sensors}} 
\author{Ranaraja S.V.\\University of Moratuwa\\Email: ranarajasv.20@uom.lk}
\date{\today}  % Date can be customized as needed

\begin{document}

\maketitle

\begin{abstract}
Navigating dynamic and unstructured environments is one of the most challenging issues in robotics. Specifically, understanding how occluded areas impact effective navigation, due to the uncertainty they introduce, is crucial. In many poorly mapped areas, conventional sensors often fail to detect obstacles until they are dangerously close. This is especially true in crowded environments, where human movement and other physical barriers frequently cause occlusions for the robot.

We propose a novel technique for inferring para-occluded obstacles by leveraging human behavioral patterns. This approach combines sensor fusion, historical trajectory data, and predictive modeling to anticipate potential obstacle locations, along with timestamps indicating when these areas are likely to be occupied. Since humans naturally avoid these areas, timely obstacle prediction allows robots to proactively adjust their paths, enhancing their ability to handle dynamic obstacles with lower collision risks and greater navigation efficiency.

The framework was validated through extensive simulations and real-world experiments, yielding significant improvements in occlusion-aware obstacle prediction. The results highlight the substantial benefits of occlusion-aware predictions in enhancing the safety and flexibility of autonomous robots operating in complex, dynamic environments.
\end{abstract}

% ====== KEYWORDS ======

\textbf{GitHub:} 
\href{https://github.com/sithija-vihanga/occlusion-aware-obstacle-prediction}{occlusion-aware-obstacle-prediction}

\begin{figure*}[ht]
    \centering
    \includegraphics[width=\textwidth]{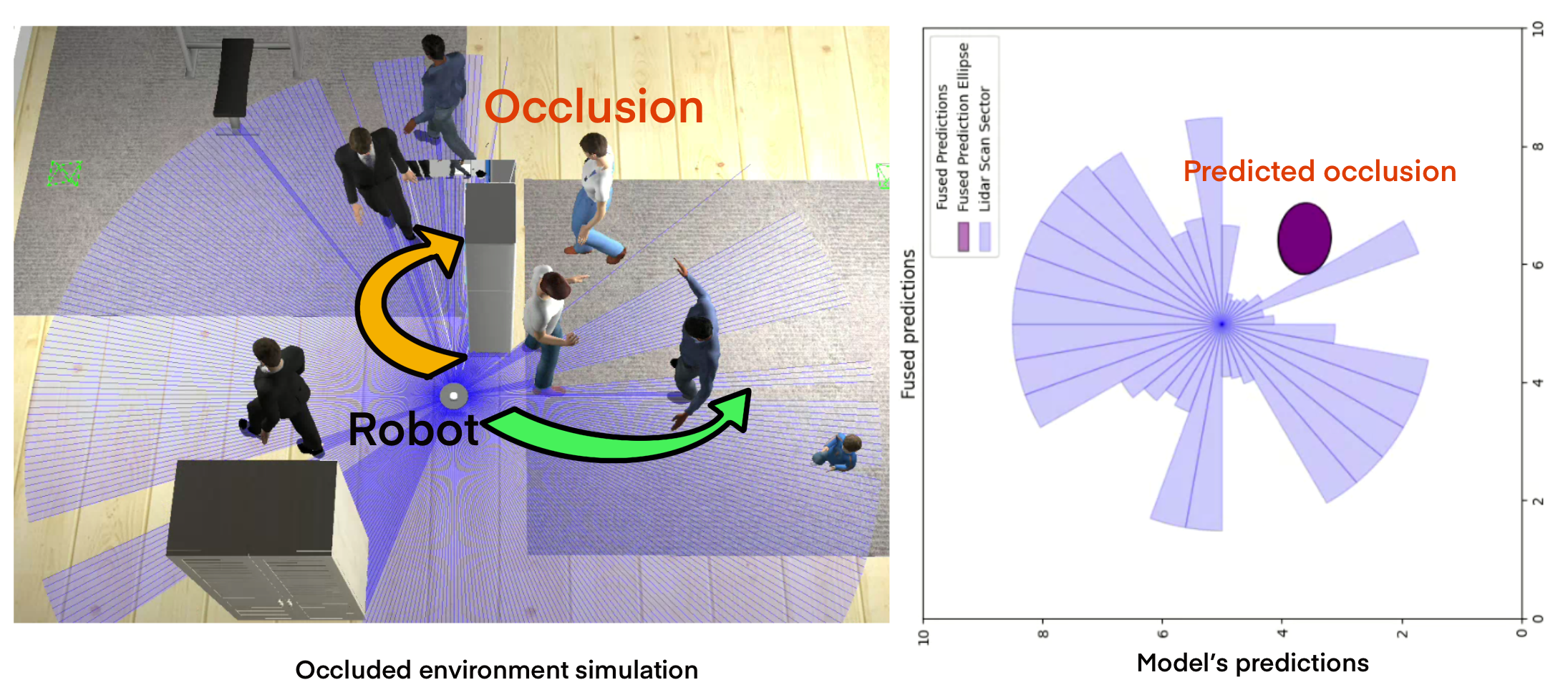}
    \caption{Occlusion handling in navigation}
    \captionsetup{font=small}
    \caption*{Robot needs insights about occlusions for path planning. Selecting collision trajectory without occlusion data (Yellow path) as it gives shorter path or selecting longer path for collision avoidance by analyzing occlusion predictions. Model predicts occlusions, human agents, static obstacles and also keeps previously detected regions for given period of time for collision avoidance.}
    \label{fig:top-image}
\end{figure*}

% ====== MAIN SECTIONS ======
\section{Introduction}

Crowd navigation is a vital area in robotics, where robots must efficiently and safely maneuver through dynamic environments populated by human agents. Many existing algorithms, such as Social Force \cite{helbing1995social}, Social GAN \cite{gupta2018social}, ORCA \cite{alonso2013optimal}, and Velocity Obstacles, assume full visibility of the environment; however, real-world situations frequently involve obstructions along the robot’s line of sight, creating considerable challenges in perceiving the environment. Occlusion-aware prediction aims to enhance accuracy in settings with restricted visibility by determining the occupancy of hidden areas that may contain either static or moving obstacles. Humans naturally rely on their past experiences to gain insight into occluded regions, a capability that inspires robotic approaches to emulate similar predictive behaviors. However, conventional path prediction approaches often fail in crowded environments due to occluded areas and the sudden changes in motion patterns of human agents, presenting additional challenges for navigation. The fast-paced advancement of assistive humanoid robots requires reliable real-time solutions to navigate through crowds, making occlusion-aware obstacle prediction increasingly important. Improving the balance between efficiency and accuracy is key to helping robots perform smoothly in highly dynamic environments. Our research focuses on leveraging human-agent interactions to provide robots with better perception and navigation skills, especially in complex and dynamic situations.

\section{Related Work}

Real-time performance is essential for enabling mobile robots to predict and respond to obstacles in occluded environments. However, there is a gap between achieving real-time performance and maintaining high accuracy in occlusion-aware prediction models. Lightweight architectures, such as MobileNet \cite{chilukuri2022robust} and YOLO \cite{9303478}, typically handle partially occluded situations using vision-based techniques but struggle in fully occluded settings. Some researchers employ image generation methods, like GANs and semantic segmentation-based inpainting algorithms \cite{lu2020semantic}, \cite{purkait2019seeing}, to infer obstacles from occluded map details, though these methods often fail to provide timely predictions. Neural network-based techniques, as outlined in \cite{ebert2017self}, \cite{dequaire2018deep}, and \cite{wang2021learning}, offer more precise occlusion predictions by leveraging prior data, but they are computationally expensive, limiting their application to smaller mobile robots. These challenges highlight the trade-off between accuracy and real-time performance.

Data-driven methods include U-Net-based models \cite{katyal2019uncertainty}, RL models \cite{isele2018navigating}, and transformer-based approaches \cite{mun2023occlusion} that estimate occlusions by encoding occluded map details and decoders trained on fully visible maps. These approaches offer generally better frameworks for occlusion prediction, but their high computational demands obstruct real-time use. Other approaches \cite{amirian2021we} use geometric crowd flow and clustering techniques to predict occlusions, but the increasing number of agents significantly increases computational requirements, especially when considering human-human interactions. 

Real-time occlusion prediction is vital for effective path planning and collision avoidance. Transformer models \cite{mun2023occlusion} and neural network-based methods \cite{tordesillas2019faster} demonstrate the potential of occlusion prediction to enhance navigation. Additionally, some studies focus on the statistical behavior of visible human agents to anticipate occlusions based on their movements \cite{hara2020predicting}, \cite{afolabipeople}. Our research extends this by considering human agents as dynamic sensors, effectively broadening the field of view (FOV) of conventional robotic sensors.

\section{Problem Statement}

Recent methods for predicting occlusions in dynamic environments face significant challenges in practical applications, particularly for mobile robots. Existing models struggle to accurately predict the positions of occluded agents, especially in densely packed settings, leading to uncertainties that negatively impact navigation and decision-making in such environments.

Most data-driven approaches rely heavily on large datasets to train models. However, creating a comprehensive dataset that covers all possible occlusion scenarios is a challenging task. The diversity of environments, agent behaviors, and interactions makes it nearly impossible to generate exhaustive data that captures the full complexity of occlusions. Additionally, many researchers argue that these systems are computationally expensive and impractical for real-time use on mobile robots, which have limited processing power. The significant time and resources required by such systems make them inefficient in fast-paced, dynamic environments.

Current models also struggle to maintain accuracy in unfamiliar or new environments, especially when dealing with densely packed human agents. As human-agent interactions become more complex in crowded spaces, existing methods fail to provide reliable predictions.

We are developing a new occlusion prediction model that enhances accuracy as the number of human agents increases. By incorporating timestamps, robots will be able to predict when occlusions are likely to be occupied and adjust their behavior accordingly to navigate more effectively in crowded spaces.

\section{Methodology}

In our approach, we focus on predicting occlusions by analyzing the reactive behaviors of observable human agents. Each predicted occlusion is validated by considering the movements of other agents in the area, utilizing a Kalman filter-based sensor fusion to reduce uncertainty and improve accuracy as the number of agents increases. To address mispredictions, we introduce a sensor-based clearing module that refines predictions by eliminating areas with lower occlusion probabilities. Our methodology combines a noisy occlusion prediction model, sensor-based clearing, and Kalman filter-based fusion to refine the prediction outputs.

\subsection{Noisy Prediction Model}

This module consists of a simplified mathematical model for predicting obstacles based on human agents' navigation patterns. It is activated by sudden movements of human agents, assuming they follow a reaction-based navigation approach in highly crowded environments. The model is triggered by thresholding two variables: the average turning angle and the maximum turning angle, both derived from dynamic windows that observe 1-second segments of the human agents' trajectory. The model predicts obstacles in three regions: in front, to the right, and to the left of the agent, with the assumption that these obstacles reflect from sudden changes in the agents' trajectories.

\begin{figure}[t]
    \centering
    % First image
    \begin{minipage}{0.5\textwidth}
        \centering
        \includegraphics[width=\linewidth]{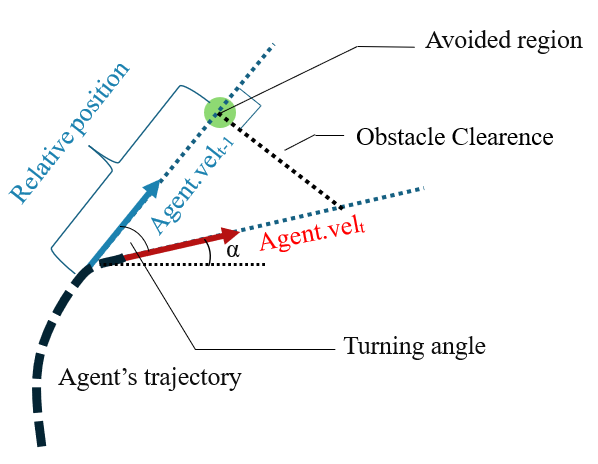}
        \caption*{(a) Front obstacle estimation through avoided regions of the human agent}
        \label{fig:calc01}
    \end{minipage}
    \hfill
    % Second image
    \begin{minipage}{0.5\textwidth}
        \centering
        \includegraphics[width=\linewidth]{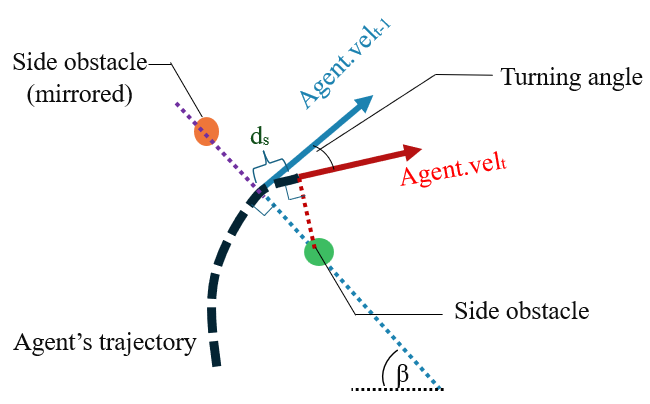}
        \caption*{(b) Side obstacle estimation by center of instantaneous velocity and its mirrored position perpendicular to agents motion}
        \label{fig:calc02}
    \end{minipage}

    \caption{Noisy obstacle estimation for front and side obstacles upon triggered by detection of sudden change in motion}
    \label{fig:calc_combined}
\end{figure}

The calculation of front obstacles takes into account the regions avoided by the human agent, maintaining a minimal social distance (obstacle clearance) according to the environmental type, as shown in \ref{fig:calc_combined}. For side obstacles, as depicted in \ref{fig:calc_combined}, the module uses the instantaneous center of zero velocity within a dynamic window and its mirrored counterpart to predict potential regions based on the human's trajectory.

\textbf{For front-obstacles:}
\begin{equation}
  \text{relative\_position} = \left| \frac{\text{obstacle\_clearance}}{\tan(\text{turning\_angle}) + \epsilon} \right|
\end{equation}

\begin{equation}
  \alpha = \tan^{-1} \left( \frac{\text{agent\_vel.y}}{\text{agent\_vel.x} + \epsilon} \right)
\end{equation}

\textbf{For side-obstacles:}
\begin{equation}
  d_S = \sqrt{ 
    \left( \text{pos.x}_{\text{t}} - \text{pos.x}_{\text{t-1}} \right)^2 
    + 
    \left( \text{pos.y}_{\text{t}} - \text{pos.y}_{\text{t-1}} \right)^2 
  }
\end{equation}
\begin{equation}
  \text{radius} = \frac{d_S}{\text{turning\_angle} + \epsilon}
\end{equation}

\begin{equation}
  \beta = \tan^{-1} \left( \frac{\text{-agent.vel.x}}{\text{agent.vel.y} + \epsilon} \right)
\end{equation}

\begin{figure}[t!]
    \centering
    \begin{minipage}[t]{0.49\textwidth}
        \centering
        \includegraphics[width=\textwidth, keepaspectratio]{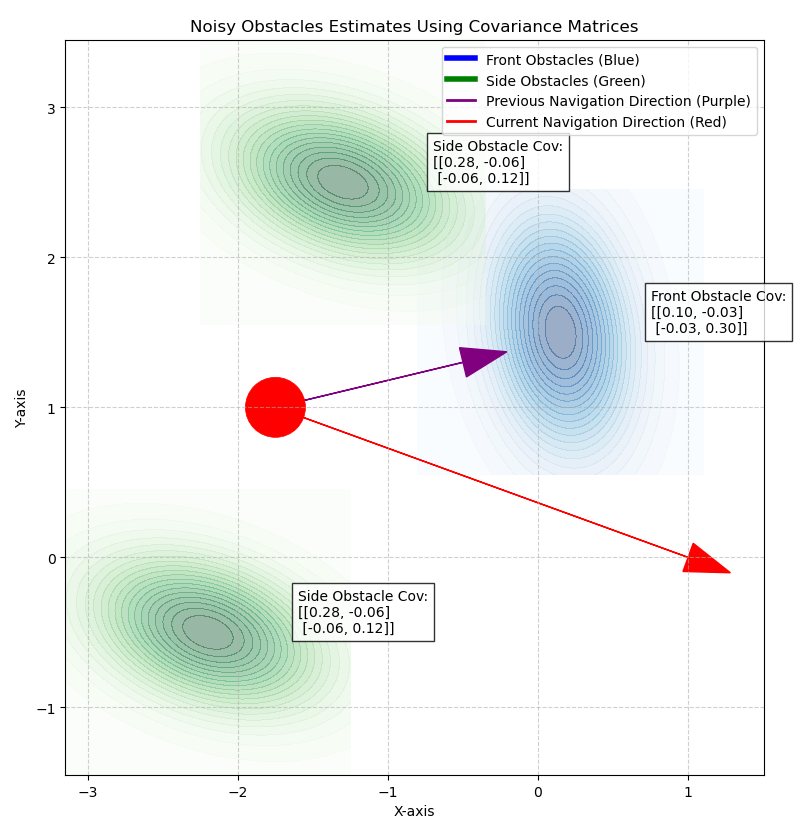}
        \caption{Gaussian obstacle prediction based on agents' moving direction and clock-wise turning angle}
        \captionsetup{font=small}
        \caption*{Side obstacles (Green) elongated along current moving direction and front obstacle (Blue) place opposing previous moving direction and aligned with current direction considering agent as expert of obstacle avoidance}
                \label{fig:gauss001}
    \end{minipage}
    \hfill
    \begin{minipage}[t]{0.49\textwidth}
        \centering
        \includegraphics[width=\textwidth, keepaspectratio]{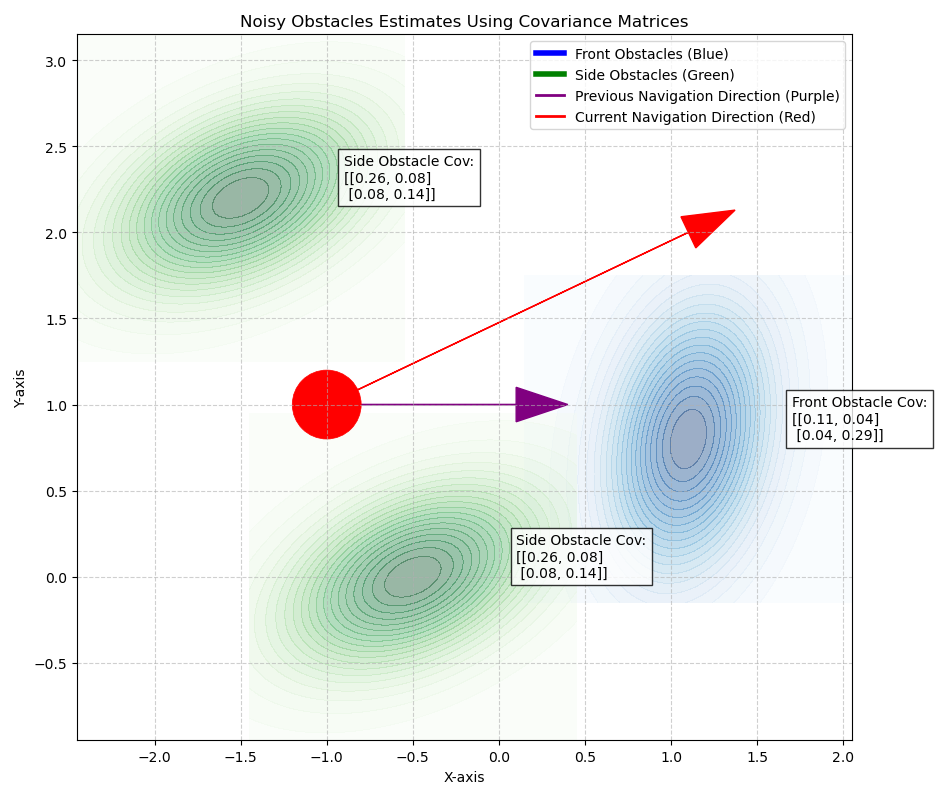}
        \caption{Gaussian obstacle prediction based on agents' moving direction and anti-clock-wise turning angle}  
        \captionsetup{font=small}
        \caption*{Side obstacles (Green) elongated along current moving direction and front obstacle (Blue) place opposing previous moving direction and aligned with current direction considering agent as expert of obstacle avoidance}
        \label{fig:gauss002}
    \end{minipage}
\end{figure}

Each predicted point corresponds to a 2D Gaussian distribution based on the agent's velocity and turning angle, accounting for the uncertainty in the prediction. The figures \ref{fig:gauss001}, \ref{fig:gauss002} illustrate the covariance matrix used to generate the obstacle regions based on turning angle. (Clockwise or anti-clockwise)

The covariance matrix calculation \ref{eq:covariance_matrix} for the Gaussian patch representing obstacles depends on the \textbf{current moving direction} \ref{eq:updated_gradient} and \textbf{average speed} of the human agent and incorporates adjustments for \textbf{the turning angle}\ref{eq:rotation_matrix_definition}. The initial covariance matrix \ref{eq:covariance_matrix} is based on the agent's speed to adjust the uncertainty, and the matrix is then rotated to align with the agent's current direction. Obstacle regions on the sides elongate along the moving direction, while the front obstacle is positioned opposing to the moving direction, considering the turning angle \ref{fig:gauss001}, \ref{fig:gauss002}. These obstacles represent potential areas that could influence the reactive motion of the human agent.

The gradient of agent's moving direction \( m \) is computed based on the velocity of the human agent at the current (\( t \)) and previous (\( t-1 \)) time steps:
   \begin{equation}
   m = 
   \begin{cases} 
     \frac{v_{y_t} - v_{y_{t-1}}}{v_{x_t} - v_{x_{t-1}}}, & \text{if} \quad v_{x_t} \neq v_{x_{t-1}} \\ 
     \infty, & \text{if} \quad v_{x_t} = v_{x_{t-1}} 
   \end{cases}
   \label{eq:updated_gradient}
   \end{equation}

The initial covariance matrix of the predicted obstacle is defined based on the average speed (\( \text{avg\_speed} \)):
\begin{equation}
\mathbf{C} = 
\begin{bmatrix}
\text{thresh} & \frac{\text{thresh}}{2} \\ 
\frac{\text{thresh}}{2} & \text{thresh}
\end{bmatrix}.
\label{eq:covariance_matrix}
\end{equation}

Where, 
\begin{equation}
\text{thresh} = 
\begin{cases} 
1.5 \cdot \text{avg\_speed}, & \text{if } \text{avg\_speed} < 0.2,\\ 
0.3, & \text{otherwise}.
\end{cases}
\label{eq:threshold_definition} 
\end{equation}

A rotation matrix \( \mathbf{R} \) is applied to orient the covariance matrix based on agent's moving direction, incorporating the turning angle (\( \theta_{\text{turning}} \)):
   \begin{equation}
\mathbf{R} = 
\begin{bmatrix}
\cos(\theta) & -\sin(\theta) \\ 
\sin(\theta) & \cos(\theta)
\end{bmatrix}.
\label{eq:rotation_matrix_definition}
\end{equation}
\begin{equation}
\theta = 
\begin{cases} 
\tan^{-1}(m) + \frac{\pi}{4} - \theta_{\text{turning}}, & \text{for front obstacles}, \\ 
\tan^{-1}(m) - \frac{\pi}{4}, & \text{for side obstacles}.
\end{cases}
\label{eq:angle_definition} \\
\end{equation}

The initial covariance matrix in \ref{eq:covariance_matrix} defines a Gaussian patch elongated along the \( y = x \) axis keeping and offset of \( \frac{\pi}{4} \). For the side obstacles, we subtract an angle of \( \frac{\pi}{4} \) and rotate the covariance matrix based on the agent's moving direction (\(\tan^{-1}(m)\)) \ref{eq:angle_definition}. For front obstacles We add an angle of \( \frac{\pi}{4} \) to make obstacle perpendicular to agent's motion and subtract \(\theta_{\text{turning}}\) to oppose the previous moving direction \ref{eq:angle_definition}.

The covariance matrix is rotated using:
   \begin{equation}
   \mathbf{C}_{\text{rotated}} = \mathbf{R} \mathbf{C} \mathbf{R}^\top.
   \label{eq:rotated_covariance}
   \end{equation}

This approach dynamically adjusts the covariance matrix to accurately reflect the potential obstacle regions based on the agents' trajectory and sudden behaviors with use of \ref{eq:rotated_covariance}.

Even though the noisy obstacle predictor responds to sudden changes in the behavioral patterns of the agent, it also attempts to predict obstacles based on abrupt changes in the agent's thoughts or plans, even in the absence of actual obstacles. In contrast, our approach addresses this issue by employing two clearing strategies specifically designed for this purpose: sensor-based clearing and a Kalman filter-based fusion approach. These methods help reduce uncertainty in the prediction.

Each predicted obstacle contains a timestamp value \ref{eq:timestamp}, indicating when the potential region will be occupied and decayed over time to avoid robot freezing conditions.

\begin{equation}
    \text{Timestamp} = 
    \begin{cases} 
        \frac{\text{Relative position}}{\text{Average Speed}}, & \text{for front obstacles}, \\ 
        0, & \text{for side obstacles}.
    \end{cases}
    \label{eq:timestamp}
\end{equation}

For front obstacles, the model predicts the time taken by the agent to reach the avoided regions with his average speed based on the assumption his sudden change of motion reflects obstacle avoidance. (Sudden changes of motion without reflecting obstacles will be cleared by clearing modules for minimizing the inaccurate predictions.) For side obstacles, the model predicts a 0 timestamp for the immediate nature of the obstacles. (Mostly for static obstacles)

\subsection{Sensor-Based Clearing}
The Sensor-Based Clearing module refines obstacle position estimates using real-time LiDAR data. The primary goal is to reduce false positives for potential obstacles in occluded areas and prevent navigation problems, such as robot freezing.

To achieve this, we divide the LiDAR sensor’s $360^\circ$ field of view into discrete sectors, with each sector covered by a sufficient number of rays to track a human agent. To smooth out sensor readings and reduce noise, each sector computes a \textbf{dynamic threshold} by averaging the values of neighboring rays. The \textbf{Ray value} in each sector represents the average distance to obstacles within that sector. The dynamic threshold for each sector is computed as follows:

\begin{equation}
\text{Threshold}[i] = \frac{\text{Ray}[i-1] + \text{Ray}[i] + \text{Ray}[i+1]}{3},
\end{equation}

This technique minimizes noisy fluctuations in the LiDAR data. The clearing module then filters each obstacle estimate based on its \textbf{distance from the robot} and \textbf{angle} relative to the robot’s position, compared against the dynamic threshold value for each sector.

The clearing module retains the obstacle only if its distance and angle satisfy the dynamic threshold for the corresponding LiDAR sector.
 \begin{equation} d \geq \textbf{Threshold}[\theta_{\textbf{obs}}] \quad \text{and} \quad d < \textbf{MAXRAY}, \end{equation} where, $\\\text{MAXRAY}$ represents the maximum sensor range.
\(\theta_{\textbf{obs}}\)\ represents the angle.\\
\textbf{d} is distance from the robot to obstacle.\\
\vspace{0.3cm}
This module is capable of clearing inaccurate estimates by comparing against lidar data. More importantly, without following conventional visible range clearing through lidar data, module performs specialized threshold for occlusion handling which enables obstacle predictions within visible regions which will be occupied by agents coming from occluded regions and clearing beyond closer range obstacles if consecutive sectors detect obstacle at longer range as consecutive sector cannot predict occlusions without having closer human agents to occluded area.  

 \begin{figure}[H] 
    \centering
    \includegraphics[width=\linewidth]{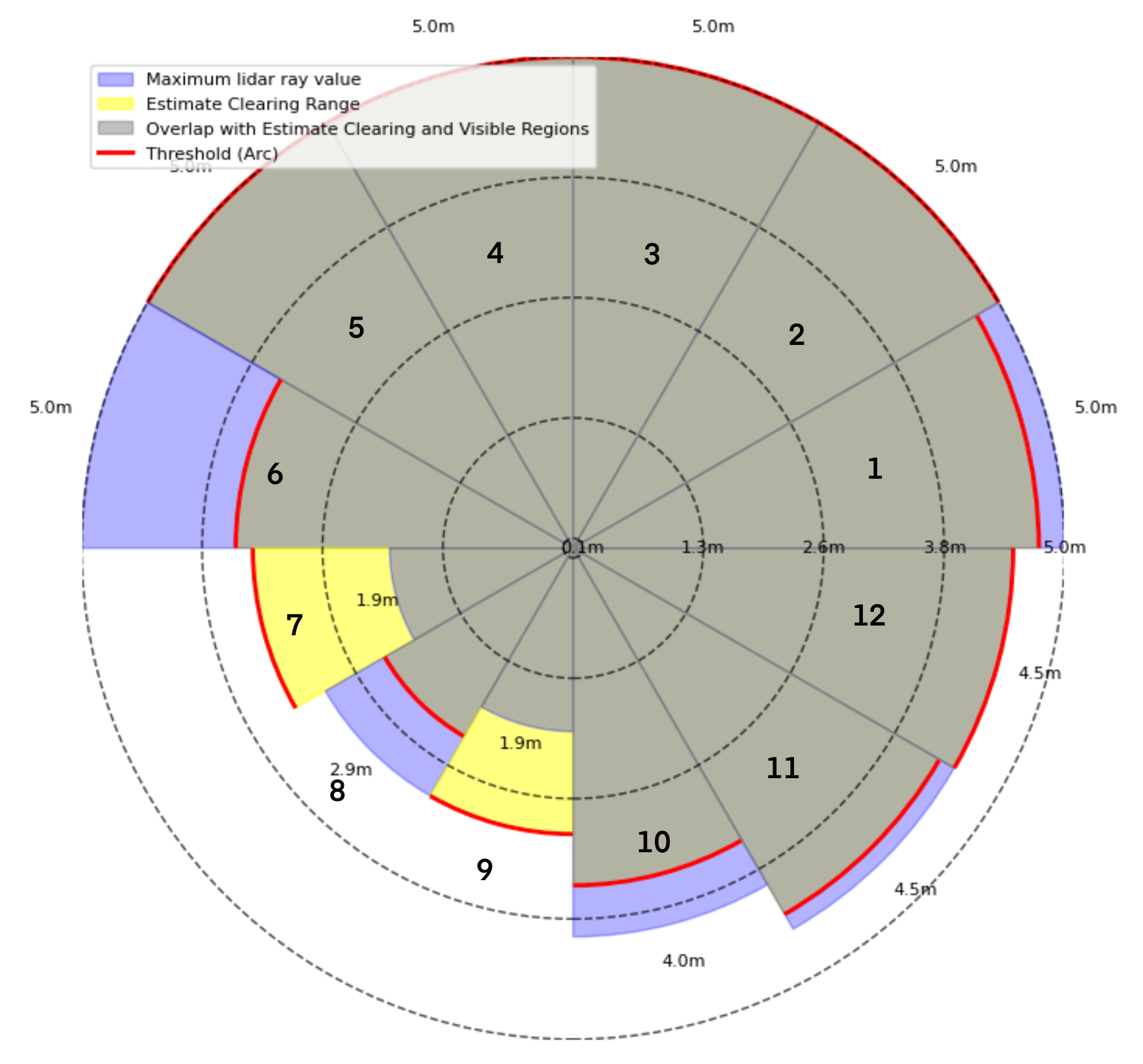}
    \caption{Sensor-based clearing sample configuration}
    \captionsetup{font=small}
    \caption*{
             Module clear all the obstacles within the estimate clearing range (Yellow or Gray). Estimated obstacles beyond the threshold and max-sensor range will be marked as obstacles. This approach enables obstacle prediction in future time steps within fully visible sectors (Obstacles coming from occluded to visible regions) [Sector 8 to 9], if any consecutive sector detects obstacles. Similarly for very close range obstacle detection, module performs clearing beyond the obstacle if other consecutive sectors detects obstacles at longer range to avoid noisy estimates. [Sector 7] }
    \label{fig:clearing}
\end{figure}

\subsection{Kalman Filter-Based Fusion}

Fusion module uses a Kalman filter for fusing the noisy predicted data from each of the human agents, thereby reducing uncertainty in the predictions. The Kalman filter is initialized after receiving the first prediction from the noisy obstacle estimator model and after clearing the data with sensor input. Each obstacle is represented as a Gaussian distribution in a 2D coordinate system. Thus, each obstacle includes the x and y positions, as well as the covariance matrix $c_{xx}$,$c_{xy}$, $c_{yx}$, $c_{yy}$, which describes the shape of the distribution, and a timestamp that indicates when the region will be occupied.

The state vector $\mathbf{x}$ is defined as:
\[
\mathbf{x}^T = \begin{bmatrix}
    dt & c_{xx} & c_{xy} & c_{yx} & c_{yy} & x & y
\end{bmatrix}
\]
\textbf{where:}  $\\$
\( dt \) is the timestamp,  $\\$
\( c_{xx}, c_{xy}, c_{yx}, c_{yy} \) are the covariance terms,  $\\$
\( x \) and \( y \) represent the position in the 2D space. $\\$

Each sensor estimate includes a timestamp, which indicates when the predicted region is expected to be occupied by an obstacle. The process noise covariance is scaled at each step according to the timestamp, accounting for the increased uncertainty in predictions with longer period for potential occupancy, and conversely, reducing uncertainty for predictions with shorter timestamps.

In the prediction step of the Kalman filter, uncertainty is added to the state without altering the position variables, as the state of an unseen obstacle cannot be predicted accurately with the available state variables. However, introducing noise based on the timestamp improves the fusion process with other data, leading to a more reliable estimate with reduced uncertainty.

The correction step involves a data extraction module that performs a nearest neighbor search to collect estimates near the predicted state (within approximately 1 meter). This module is built using a KD-tree structure to accelerate the nearest neighbor search process. These estimates are then used for sensor fusion, resulting in more accurate predictions with lower uncertainty. As each estimate is fetched, its measurement noise covariance is updated based on its corresponding timestamp.
\vspace{0.2cm}

\textbf{Covariance update based on timestamp:}

\begin{equation}
\Delta t = \texttt{timeStamp} - \texttt{timeNow}
\end{equation}
\begin{equation}
\texttt{scale} = C_1 - e^{-C_2 (\Delta t)^2}
\end{equation}
\begin{equation}
\texttt{Covariance}_{\texttt{updated}} = \texttt{scale} \cdot \texttt{Covariance}
\end{equation}

\noindent
\textbf{Constants:}
\begin{flalign*}
C_1 &= 1.1 \quad \text{(Constant offset for scale calculation)} \\
C_2 &= 0.3 \quad \text{(Exponential decay rate)}
\end{flalign*}

\begin{figure}[t]
    \centering
    \includegraphics[width=1.0\linewidth]{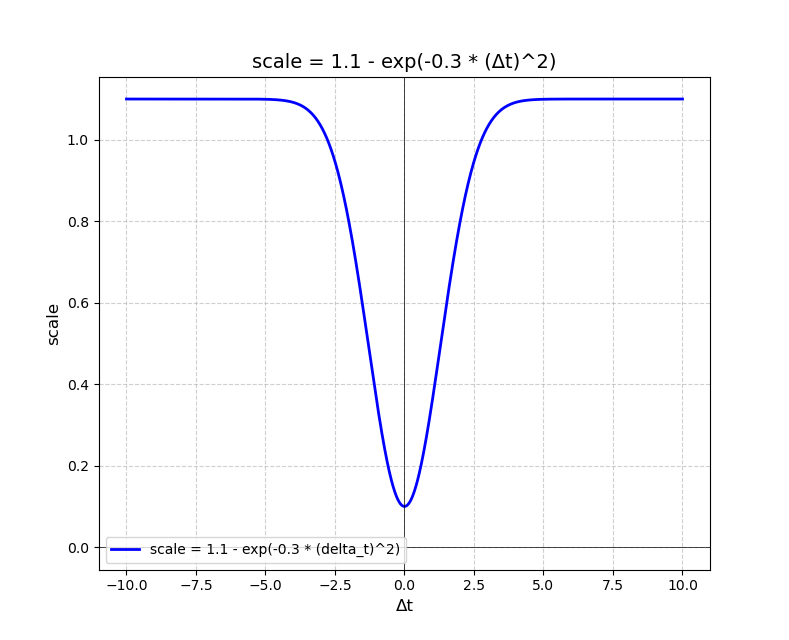}
    \caption{Scaler for covariance matrices \\
             Based on this scaling factor, any obstacle with higher timestamp will get higher uncertainity in all state variables and gets lower uncertainity when timestep is lower.}
    \label{fig:covariance_scaler}
\end{figure}

Scaler function ensures the covariance matrices of obstacles with higher $\Delta t$ to have larger uncertainity and obstacles with lower $\Delta t$ values to have lower uncertainity improving the accuracy of the predictions.

\begin{table*}[htbp]
\centering
\begin{tabular}{|c|c|c|c|c|}
\hline
\textbf{Distance Range (m)} & \textbf{Agent (\%)} & \textbf{Obstacle (\%)} & \textbf{Incorrect (\%)} & \textbf{Unseen (\%)} \\
\hline
0.0-0.5 & 0.0 & 33.33 & 0.0 & 66.67 \\
\hline
0.5-1.0 & 3.03 & 83.33 & 12.12 & 1.52 \\
\hline
1.0-1.5 & 22.47 & 32.58 & 0.0 & 44.94 \\
\hline
1.5-2.0 & 18.56 & 35.33 & 1.20 & 44.91 \\
\hline
2.0-2.5 & 16.46 & 17.09 & 11.71 & 54.75 \\
\hline
2.5-3.0 & 21.43 & 5.0 & 10.0 & 63.57 \\
\hline
\end{tabular}
\caption{Prediction Percentages by Distance Range}
\captionsetup{font=small}
\caption*{This table summarizes the predictions in each of the categorizes against the distance from the robot. Data shows overall high occlusion prediction percentage, and lower inaccurate predictions when closer to the robot. (Max. field of view used here is 3m)}
\label{tab:predictions}
\end{table*}

\section{Results \& Evaluation}

The simulation results and evaluation are based on two types of test cases:
\begin{itemize}
    \item A reaction-based human navigation simulator for model prediction testing.
    \item A social force model-based Gazebo simulator for real-world deployment testing.
\end{itemize}

The first test employs a reaction-based human navigation simulation with random obstacles and goals. It involves 80 random initializations of agent start positions, goals, and static obstacles, with the robot placed in a fixed position. Key performance metrics recorded include the accuracy of occlusion predictions, obstacle predictions, agent predictions, and the accuracy of the predicted cost (based on the predicted timestamp). Additionally, the prediction percentage across each region was analyzed in comparison to the sensor's maximum field of view. \ref{tab:predictions}

\begin{figure*}[h]
    \centering
    \includegraphics[width=\textwidth]{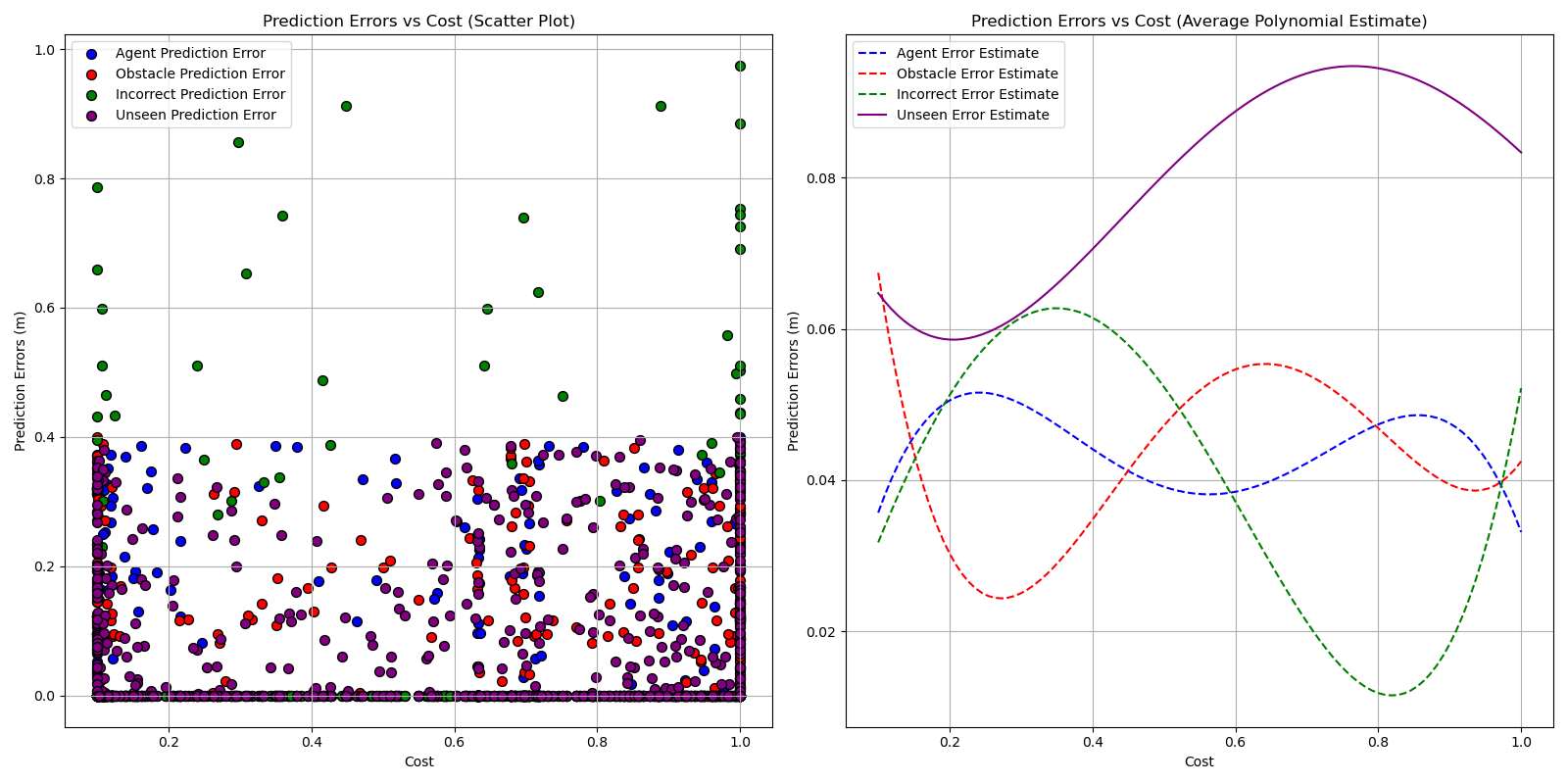}
    \caption{Model performance: Predictions vs Cost.}
    \captionsetup{font=small}
    \caption*{As shown on the left side of Figure, many predictions fall within a 40 cm error margin. Occlusion predictions are saturated at cost values of 0.1 (static agents) and 1 (static and dynamic agents), reflecting their relative threat levels. Having significant number of occlusion predictions over 0.6 cost value indicates the validation of estimates through the timestamps (cost is inversely proportional to timestamp)

On the right, the average prediction error for occlusions is around 10 cm at higher cost values (when estimating dynamic agents) by reducing the inaccurate predictions show the reliability of the model. At lower cost values occlusion predictions have lower error due to estimating static agents.}
    \label{fig:predictionvscost}
\end{figure*}

The performance evaluation was based on 1,780 predictions, which included occlusions, agents, and obstacle predictions over 80 random initial environmental setups. As shown in the scatter plot \ref{fig:predictionvscost}, occlusion predictions are evaluated by analyzing the distance between a given prediction and closest static obstacle or human agent, referred to as \textbf{prediction error}. A prediction is considered incorrect if the error exceeds 40 cm and each obstacle considered to have a footprint of 40 cm during the simulation. Human agent or obstacle predictions are categorized based on whether the model predicts a human agent or a static obstacle. 

As shown in the left side of the \ref{fig:predictionvscost} scatter plot, many predictions are concentrated within a 40 cm error margin. Additionally, occlusion predictions are saturated at both ends of the cost scale (inversely proportional to timestamp), with cost values of 0.1 and 1. This corresponds to static and dynamic agents with immediate occupancy in the predicted region (cost = 1) and static agents in the region after exceeding the timestamp (cost = 0.1) as timestamp (cost) decays over time to avoid conflicts. A significant number of occlusions appear above the 0.6 cost value indicating how timestamp (cost) can validate the dynamic occlusions. The model focuses more on dynamic occlusion predictions, as static agents do not pose a significant threat in occlusion-aware navigation.

In the average prediction error estimate shown on the right side of \ref{fig:predictionvscost}, the average prediction error for occlusion predictions is around 8 cm. Furthermore, agent and obstacle predictions exhibit lower prediction errors compared to occlusion predictions. However, this level of occlusion prediction is sufficient for crowd navigation involving occlusions.

Model performance was tested by evaluating the distance between the robot and the predicted points. The model’s prediction range extends from the robot to the maximum sensor field of view (maximum lidar range). In the reaction-based simulator, the maximum field of view is set to 3 meters. The table \ref{tab:predictions} summarizes the prediction accuracy in each region up to the maximum sensor range (3 meters). The model performs best in the 0-0.5m range with 66.67\% accuracy which directly relates to occlusions impacting with the robot. Agent predictions are excluded in this range, as only cross-agent predictions are considered, leading to a 0.0\% agent prediction rate in closer ranges due having agent footprint of 40 cm. Incorrect predictions are also minimal in this range due to having closer obstacles which leads to occlusions. For the region between 0.5-1.0 m; model predicts obstacles which creates the occlusions and for 1.0-1.5 m predictions indicate unseen obstacles with lower inaccurate predictions due to the performance for clearing module in occluded areas. For other regions, incorrect predictions increase, but on average, the model performs best in occlusion predictions. So the model performs its best up to half of max sensor range (up to 1.5 m) as clearing module fails at max sensor range.

The second test involved integrating the algorithm into the social force model plugin-based crowd simulator \cite{socialforcemodel-repo}, using a TurtleBot3 Waffle robot to obtain lidar readings. Human detection and tracking were performed using a leg detector package \cite{legdetector-repo} based on the lidar readings. The simulation was conducted with 8 human agents, incorporating occlusions by adding static obstacles and stationary human agents. The model successfully predicted cost values and occlusion regions in real time, validating the results from the first test with the reaction-based simulator.

\section{Conclusion}
\vspace{0.5cm}

The suggested architecture for real-time obstacle prediction in occluded spaces is intended to improve environmental perception for robot path planning and navigation in crowded settings. Initially, all possible obstacle regions are considered, and then the model employs a probabilistic approach to increasingly modify predictions with accurate identification of dynamic obstacles through Kalman filter-based data fusion. This method effectively handles uncertainties by manipulating 2D Gaussian distributions for predicted potential obstacle regions. 

Each potential obstacle estimate from the model yields positions with reduced uncertainty by the Kalman filter, which leads to greater accuracies with an increasing number of human agents within crowds. By validating predictions based on the actions of individual human agents, the system is assured of reliable, context-aware obstacle prediction. The clearing module maintains minimal interference from noisy or wrong prediction values, improving overall model performance. In addition, the dynamic cost values address the problem of robot freezing in tight spaces by providing real-time cost values for advanced navigation algorithms. The simulation results prove the ability of the system to accurately predict real-time obstacles in occluded regions, which increases its potential use in highly complex and dynamic environments.

\section*{Acknowledgements}
I sincerely thank Rajarathne G.K.M.I.D for his invaluable support in testing the occlusion-aware obstacle prediction framework, providing critical feedback, and ensuring its robustness and accuracy.

\bibliographystyle{plain}
\bibliography{references}

\begin{thebibliography}{10}

\bibitem{socialforcemodel-repo}
lightsfm.
\newblock \url{hhttps://github.com/robotics-upo/lightsfm}, 2016.
\newblock Accessed: 2024-12-28.

\bibitem{legdetector-repo}
ros2\_leg\_detector.
\newblock \url{https://github.com/mowito/ros2_leg_detector}, 2020.
\newblock Accessed: 2024-12-28.

\bibitem{afolabipeople}
Oladapo Afolabi, Katherine Driggs-Campbell, Roy Dong, Mykel~J Kochenderfer, and S~Shankar Sastry.
\newblock People as sensors: Imputing maps from human actions. in 2018 ieee.
\newblock In {\em RSJ International Conference on Intelligent Robots and Systems (IROS)}, pages 2342--2348.

\bibitem{alonso2013optimal}
Javier Alonso-Mora, Andreas Breitenmoser, Martin Rufli, Paul Beardsley, and Roland Siegwart.
\newblock Optimal reciprocal collision avoidance for multiple non-holonomic robots.
\newblock In {\em Distributed autonomous robotic systems: The 10th international symposium}, pages 203--216. Springer, 2013.

\bibitem{amirian2021we}
Javad Amirian, Jean-Bernard Hayet, and Julien Pettre.
\newblock What we see and what we don't see: Imputing occluded crowd structures from robot sensing.
\newblock {\em arXiv preprint arXiv:2109.08494}, 2021.

\bibitem{chilukuri2022robust}
Devi~M Chilukuri, Sun Yi, and Younho Seong.
\newblock A robust object detection system with occlusion handling for mobile devices.
\newblock {\em Computational Intelligence}, 38(4):1338--1364, 2022.

\bibitem{dequaire2018deep}
Julie Dequaire, Peter Ondr{\'u}{\v{s}}ka, Dushyant Rao, Dominic Wang, and Ingmar Posner.
\newblock Deep tracking in the wild: End-to-end tracking using recurrent neural networks.
\newblock {\em The International Journal of Robotics Research}, 37(4-5):492--512, 2018.

\bibitem{ebert2017self}
Frederik Ebert, Chelsea Finn, Alex~X Lee, and Sergey Levine.
\newblock Self-supervised visual planning with temporal skip connections.
\newblock {\em CoRL}, 12(16):23, 2017.

\bibitem{gupta2018social}
Agrim Gupta, Justin Johnson, Li~Fei-Fei, Silvio Savarese, and Alexandre Alahi.
\newblock Social gan: Socially acceptable trajectories with generative adversarial networks.
\newblock In {\em Proceedings of the IEEE conference on computer vision and pattern recognition}, pages 2255--2264, 2018.

\bibitem{hara2020predicting}
Kensho Hara, Hirokatsu Kataoka, Masaki Inaba, Kenichi Narioka, Ryusuke Hotta, and Yutaka Satoh.
\newblock Predicting vehicles appearing from blind spots based on pedestrian behaviors.
\newblock In {\em 2020 IEEE 23rd International Conference on Intelligent Transportation Systems (ITSC)}, pages 1--8. IEEE, 2020.

\bibitem{helbing1995social}
Dirk Helbing and Peter Molnar.
\newblock Social force model for pedestrian dynamics.
\newblock {\em Physical review E}, 51(5):4282, 1995.

\bibitem{isele2018navigating}
David Isele, Reza Rahimi, Akansel Cosgun, Kaushik Subramanian, and Kikuo Fujimura.
\newblock Navigating occluded intersections with autonomous vehicles using deep reinforcement learning.
\newblock In {\em 2018 IEEE international conference on robotics and automation (ICRA)}, pages 2034--2039. IEEE, 2018.

\bibitem{katyal2019uncertainty}
Kapil Katyal, Katie Popek, Chris Paxton, Phil Burlina, and Gregory~D Hager.
\newblock Uncertainty-aware occupancy map prediction using generative networks for robot navigation.
\newblock In {\em 2019 International Conference on Robotics and Automation (ICRA)}, pages 5453--5459. IEEE, 2019.

\bibitem{9303478}
Yongjun Li, Shasha Li, Haohao Du, Lijia Chen, Dongming Zhang, and Yao Li.
\newblock Yolo-acn: Focusing on small target and occluded object detection.
\newblock {\em IEEE Access}, 8:227288--227303, 2020.

\bibitem{lu2020semantic}
Chenyang Lu and Gijs Dubbelman.
\newblock Semantic foreground inpainting from weak supervision.
\newblock {\em IEEE Robotics and Automation Letters}, 5(2):1334--1341, 2020.

\bibitem{mun2023occlusion}
Ye-Ji Mun, Masha Itkina, Shuijing Liu, and Katherine Driggs-Campbell.
\newblock Occlusion-aware crowd navigation using people as sensors.
\newblock In {\em 2023 IEEE International Conference on Robotics and Automation (ICRA)}, pages 12031--12037. IEEE, 2023.

\bibitem{purkait2019seeing}
Pulak Purkait, Christopher Zach, and Ian Reid.
\newblock Seeing behind things: Extending semantic segmentation to occluded regions.
\newblock In {\em 2019 IEEE/RSJ International Conference on Intelligent Robots and Systems (IROS)}, pages 1998--2005. IEEE, 2019.

\bibitem{tordesillas2019faster}
Jesus Tordesillas, Brett~T Lopez, and Jonathan~P How.
\newblock Faster: Fast and safe trajectory planner for flights in unknown environments.
\newblock In {\em 2019 IEEE/RSJ international conference on intelligent robots and systems (IROS)}, pages 1934--1940. IEEE, 2019.

\bibitem{wang2021learning}
Lizi Wang, Hongkai Ye, Qianhao Wang, Yuman Gao, Chao Xu, and Fei Gao.
\newblock Learning-based 3d occupancy prediction for autonomous navigation in occluded environments.
\newblock In {\em 2021 IEEE/RSJ International Conference on Intelligent Robots and Systems (IROS)}, pages 4509--4516. IEEE, 2021.

\end{thebibliography}

\begin{figure*}[htbp]
    \centering
    \includegraphics[width=\textwidth]{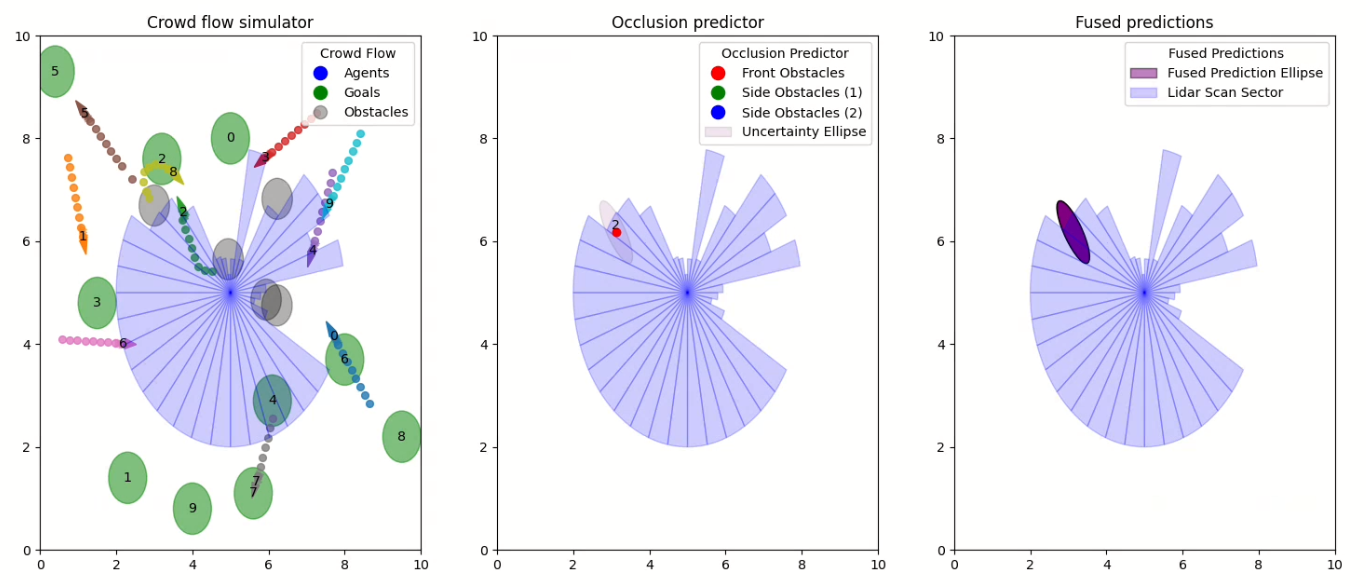}
    \caption{Model performance at time = t-$\delta$t}
    \label{fig:prediction_t_1}
\end{figure*}

\begin{figure*}[htbp]
    \centering
    \includegraphics[width=\textwidth]{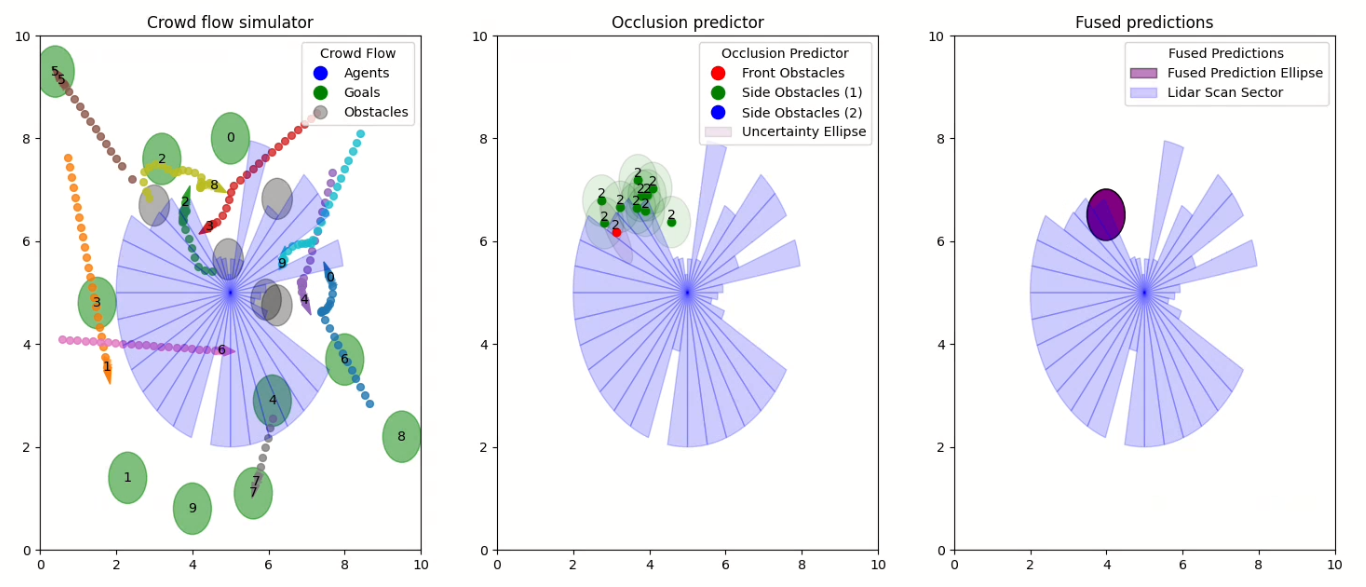}
    \caption{Model performance at time = t}
    \label{fig:prediction_t}
\end{figure*}

\begin{figure*}[htbp]
    \centering
    \includegraphics[width=\textwidth]{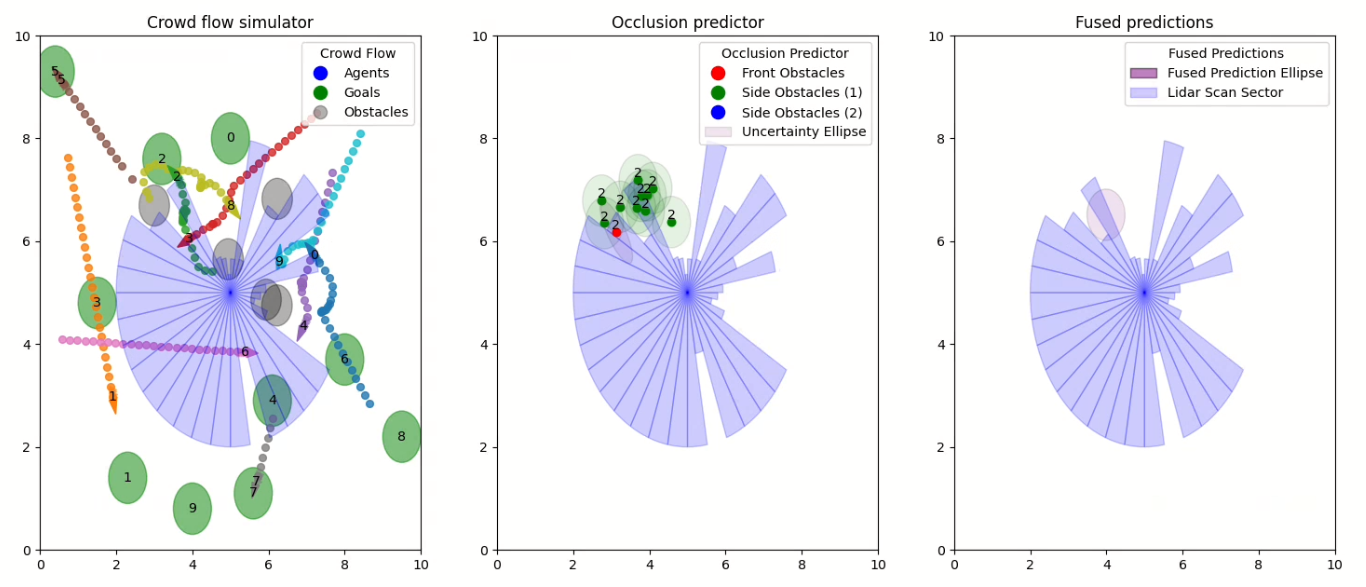}
    \caption{Model performance at time = t+$\delta$t}
    \label{fig:prediction_t1}
\end{figure*}

% ====== END DOCUMENT ======
\end{document}